\documentclass[sn-vancouver,Numbered,iicol]{sn-jnl} 

\usepackage{lineno}
 
\usepackage{graphicx} 
\usepackage{multirow} 
\usepackage{amsmath,amssymb,amsfonts} 

\usepackage{alphabeta} 
\usepackage{siunitx}	

\begin{document}


\title{Generative AI and Foundation Models in Medical Image}

\author*[1,2]{Masahiro Oda} 
\email{moda@mori.m.is.nagoya-u.ac.jp, oda.masahiro.h6@f.mail.nagoya-u.ac.jp}

\affil[1]{Data Science Research Division, Information Technology Center, Nagoya University, Furo-cho, Chikusa-ku, Nagoya 464-8601, Japan}
\affil[2]{Graduate School of Informatics, Nagoya University, Furo-cho, Chikusa-ku, Nagoya 464-8601, Japan}

\abstract{
In recent years, generative AI has attracted significant public attention, and its use has been rapidly expanding across a wide range of domains. From creative tasks such as text summarization, idea generation, and source code generation, to the streamlining of medical support tasks like diagnostic report generation and summarization, AI is now deeply involved in many areas. Today's breadth of AI applications is clearly distinct from what was seen before generative AI gained widespread recognition.
Representative generative AI services include DALL·E 3 (OpenAI, California, USA) and Stable Diffusion (Stability AI, London, England, UK) for image generation, ChatGPT (OpenAI, California, USA), and Gemini (Google, California, USA) for text generation. The rise of generative AI has been influenced by advances in deep learning models and the scaling up of data, models, and computational resources based on the scaling laws.
Moreover, the emergence of foundation models, which are trained on large-scale datasets and possess general-purpose knowledge applicable to various downstream tasks, is creating a new paradigm in AI development. These shifts brought about by generative AI and foundation models also profoundly impact medical image processing, fundamentally changing the framework for AI development in healthcare.
This paper provides an overview of diffusion models used in image generation AI and large language models (LLMs) used in text generation AI, and introduces their applications in medical support.
This paper also discusses foundation models, which are gaining attention alongside generative AI, including their construction methods and applications in the medical field.
Finally, the paper explores how to develop foundation models and high-performance AI for medical support by fully utilizing national data and computational resources.
}


\maketitle


\section{Introduction} 

In November 2022, ChatGPT (OpenAI, California, USA), powered by GPT-3.5 model, was released. Due to its significantly improved performance compared to previous chatbot AI models, it quickly gained global attention. Around the same time, Stable Diffusion emerged, enabling high-quality image generation and signaling a technological transformation in image generation. Since then, generative AI has become widely recognized, leading to numerous research projects and commercial services leveraging this technology.
Today, generative AI is used in a wide range of applications, from creative tasks such as text generation, summarization, idea generation, illustration, digital art, and source code generation to the streamlining of medical support tasks like diagnostic report generation and summarization. The breadth of generative AI applications significantly departs from the pre-generative AI era.
Notable generative AI services include DALL·E 3 (OpenAI, California, USA), Stable Diffusion (Stability AI, London, England, UK) for image generation, and ChatGPT and Gemini (Google, California, USA) for text generation. The rapid rise of generative AI has been driven by advancements in deep learning models and changes in training methodologies. These innovations can potentially reshape the framework for AI development in medical image processing.

This paper provides an overview of diffusion models and large language models (LLMs) used in generative AI development, along with examples of their applications in medical image processing. 
Additionally, this paper discusses foundational models, which have garnered attention with generative AIs, and explores new AI development approaches utilizing these models.
Finally, this paper examines strategies to leverage national computational and informational resources to develop internationally competitive, nationally produced foundational models and high-performance AI for medical image processing.

\section{Discriminative AI and generative AI}

The introduction of AlexNet \cite{Alex12} in 2012 marked the beginning of active research in image processing using deep learning. At that time, most deep learning models took data such as images as input and produced classification or detection results as output. These models can be categorized as ``Discriminative AI.'' The main tasks performed by Discriminative AI include image classification, object detection, and segmentation. Among these, segmentation is a process that performs pixel-wise classification. A key characteristic of Discriminative AI is that the output data contains less information than the input data. Additionally, models such as Convolutional Neural Networks (CNNs) and Fully Convolutional Networks (FCNs) are commonly used in its architecture.

Generative AI, on the other hand, takes inputs such as text and produces outputs such as text or images using a generation model. A defining feature of Generative AI is its ability to generate high-information outputs, such as long text or images, from relatively low-information inputs, such as short sentences. Recent Generative AI models often adopt Diffusion Models or large-scale Transformers \cite{Transformer} as their architecture, although some Generative AI models still use CNNs or FCNs.
Details of diffusion models are explained in \ref{ssec:diffusionmodel}.

A common way to interact with Generative AI is by providing instructions in the form of text, leading to the widespread use of the term ``prompt'' to describe the input given to Generative AI. In the past, due to AI's limited capabilities, it was seen as merely a machine that followed human instructions. However, with recent advancements in AI performance, Generative AI is increasingly perceived as an entity capable of intelligent reasoning, which has led to the adoption of the term ``prompt,'' like how humans give instructions to one another.

The following chapters will discuss image generation AI and text generation AI, both of which fall under the category of Generative AI.

\section{Image generation AI}

\subsection{Overview}

Previously, Variational Autoencoders (VAEs) \cite{VAE} and Generative Adversarial Networks (GANs) \cite{GAN} were commonly used for image generation. VAEs offer stable training and make it easier to evaluate training progress, but they struggle to create highly expressive models. On the other hand, GANs can generate highly detailed images, as seen in StyleGAN2 \cite{Tero15}, which has demonstrated impressive facial image generation. However, GANs suffer from unstable training and difficulty in evaluating training progress.

Recent image generation services like DALL·E 3, Stable Diffusion, and Midjourney can produce more visually appealing illustrations and photorealistic images than VAEs or GANs. These services utilize diffusion models for image generation.

This chapter discusses the mechanism of diffusion models and their applications in medical image processing.

\subsection{Diffusion models}
\label{ssec:diffusionmodel}

This section provides a brief explanation of diffusion models. Diffusion models are inspired by thermodynamics, where the concentration of substances, temperature, and energy differences tend to equalize over time. This phenomenon occurs because molecules within a substance undergo random motion due to thermal energy, gradually mixing until they achieve a uniform distribution. The Diffusion Probabilistic Model (DPM) \cite{Jascha15}, which forms the foundation of diffusion models, was developed based on this diffusion process. In DPM, noise is gradually added to the data distribution, transforming it into a completely noisy distribution (diffusion process). The model then reconstructs the data distribution by reversing this process (reverse diffusion process).
Please note that the assumption of the existence of reverse transformations of the diffusion process in DPM is different from the diffusion process in thermodynamics.

\begin{figure}[tb]
\centering
\includegraphics[width=\linewidth]{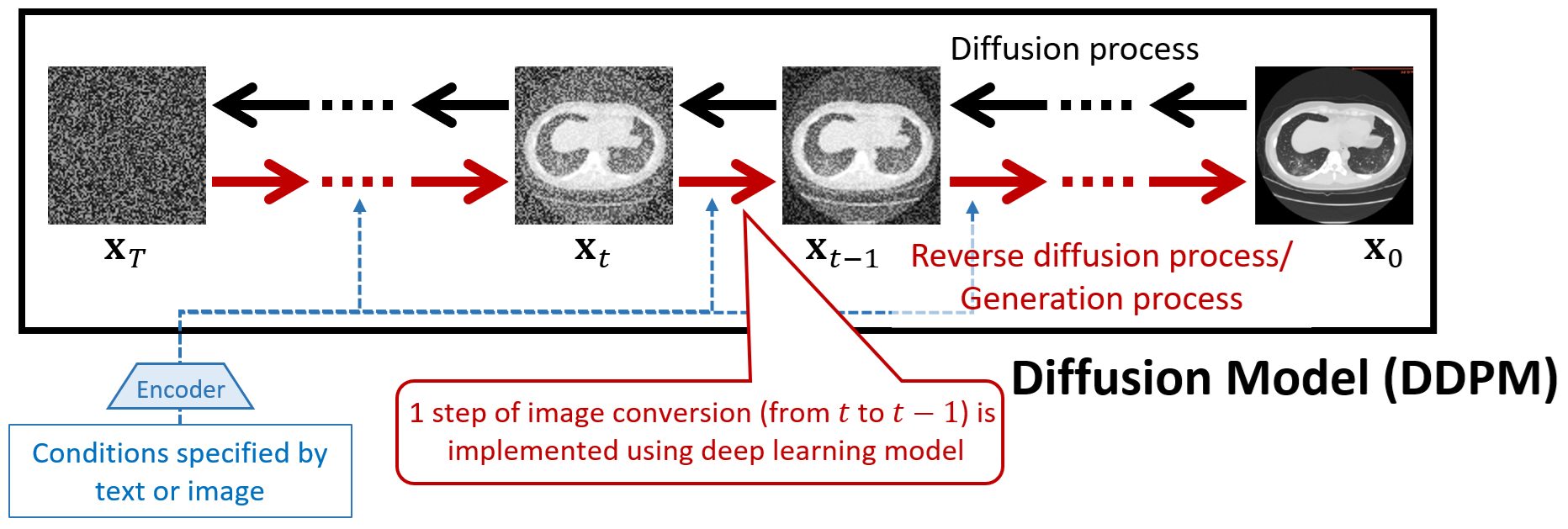}
\caption{Conceptual diagram of task-specific AI development using foundation models}
\label{fig:dm}
\end{figure}

For image generation, a commonly used variant of DPM is the Denoising Diffusion Probabilistic Model (DDPM) \cite{Jonathan20}. In DDPM, an image undergoes a diffusion process, where gradual noise is added until the image is completely transformed into pure noise. Then, through the reverse diffusion process, the model progressively removes the noise to reconstruct an image. Since this reverse diffusion process creates an image, it is also called the generation process. This process can be understood as a transformation from a noise distribution to the target image distribution. Figure \ref{fig:dm} illustrates the diffusion and reverse diffusion processes in DDPM.

In the implementation of DDPM, the reverse diffusion process, removing noise and reconstructing an image, is typically performed using a deep learning model such as U-Net \cite{U-Net}. 
The U-Net is an encoder-decoder style FCN model, which is commonly used to perform medical image segmentation.
This model takes a noisy image as input and outputs an image with slightly reduced noise. The final generated image is obtained by repeatedly applying this noise reduction process. Since the deep learning model needs to reduce noise at each iteration appropriately, it receives both the step count and generation conditions as input. The reverse diffusion process typically involves about 1,000 steps to generate high-quality images.

In image generation, text or other images are often used to control the output. This control is achieved by providing embedding representations of generation conditions to the deep learning model at each step of the reverse diffusion process. For example, in DALL·E 3, which generates images based on text descriptions, a Transformer converts text into an embedding representation, which is then used as a generation condition in the reverse diffusion process. Similarly, in diffusion model-based segmentation methods, images are converted into embedding representations and used as generation conditions.

\subsection{Medical image generation using diffusion models}

Many methods utilizing diffusion models have emerged in medical image processing, and this section, along with the following sections, introduces some of the most representative examples. Soon after diffusion models gained attention, medical image generation techniques using diffusion models were proposed.

Methods for generating 2D brain MR images \cite{Walter22}, 3D brain MR images \cite{Zolnamar22}, and 4D cardiac MR images \cite{Boah22} using diffusion models have been introduced, and these papers indicate that they achieve better results than GAN-based image generation.
Additionally, diffusion models have been applied to generate laparoscopic images during surgery \cite{Yihang25,Danush24}. These laparoscopic image generation methods introduced techniques to control the presence of surgical instruments and organs in the generated images. Zhou et al. \cite{Yihang25} used text descriptions as generation conditions, while Venkatesh et al. \cite{Danush24} used label images. In these approaches, text or label images are first converted into embedding representations, which are then fed into the diffusion model for image generation.

Here, this paper examines how generated medical images can benefit computer-aided medical applications. While diffusion models can generate realistic medical images, these generated images are not actual patient images and, therefore, cannot be used directly for diagnostic or therapeutic support. However, this does not mean that generated medical images are useless.
Image generation using diffusion models allows the incorporation of text descriptions, label images, and other generation conditions, making it possible to produce images that accurately reflect real anatomical structures, diseases, and intraoperative conditions when appropriately controlled. As a result, generated medical images play a valuable role in data augmentation for training deep learning models used in diagnostic and therapeutic support.
Compared to traditional data augmentation techniques such as geometric transformations, color transformations, and Mixup \cite{Mixup}, which have been widely used, image generation using diffusion models can create more realistic and diverse variations in datasets.

A study \cite{Abe23} explored the use of diffusion-generated images for training deep-learning models in mammogram-based diagnostic support. This study used Stable Diffusion to generate mammogram images with embedded tumors. During image generation, text-based prompts were used to control the position and size of the embedded tumors. By incorporating these generated images into the training of diagnostic support deep-learning models, the study successfully developed a higher-performance model compared to training with only real images.
Thus, generated images using diffusion models can serve as an advanced data augmentation technique, significantly enhancing the effectiveness of medical AI applications.

\subsection{Medical image segmentation using diffusion models}
\label{ssec:segmentation}

One of the most effective applications of diffusion models in medical image processing is segmentation. Among the early methods utilizing diffusion models for segmentation, MedSegDiff \cite{Junde24} was one of the first proposed approaches. This method generates segmentation result images through the reverse diffusion process, where embedding representations extracted from medical images are provided as generation conditions. This process enables the generation of segmentation results as images corresponding to the input medical images.
Compared to previously proposed FCN-based methods, such as SegNet and nnU-Net, and FCN+Vision Transformer (ViT) approaches, such as TransUNet, MedSegDiff demonstrated higher segmentation accuracy. In addition to MedSegDiff and its improved version, MedSegDiff-V2 \cite{Junde24-2}, many diffusion model-based segmentation methods have emerged and gained attention at MICCAI, a leading international conference on computer-aided medical applications.
This section introduces several diffusion model-based medical image segmentation methods.

The reverse diffusion process used for segmentation in diffusion models can be implemented using a U-Net-based model. However, U-Net may sometimes fail to generate images that faithfully correspond to the medical images given as generation conditions. To address this issue, Chowdary et al. \cite{Chowdary23} proposed a Multi-sized Transformer U-Net (MT U-Net), a deep learning model designed explicitly for diffusion model-based segmentation. In diffusion models, the deep learning model takes two inputs: the output image from the previous reverse diffusion step and the original medical image.
MT U-Net enhances image generation by extracting effective features from these two inputs using Cross Attention between image feature representations. Additionally, it incorporates a Multi-size Transformer module, which utilizes information from various spatial positions and scales for image generation. These enhancements enabled MT U-Net to achieve better segmentation results than traditional FCN-based methods, FCN+ViT-based methods, and MedSegDiff and MedSegDiff-V2.

Other segmentation approaches have improved accuracy by refining the noise-generation process in diffusion models. Typically, diffusion models add Gaussian-distributed noise to images. However, Chen et al. \cite{TaoChen23} suggested that, for segmentation tasks that involve binary images, it is more effective to use Bernoulli-distributed noise, which takes only two discrete values. Their approach, which incorporated Bernoulli noise into a diffusion model for segmentation, demonstrated superior results when applied to MRI and CT images. Their method outperformed traditional FCN-based and ViT-based approaches and diffusion model-based methods that use Gaussian noise.

Some medical images accompany text descriptions of their contents. When training segmentation models, utilizing ground-truth label images and textual information enables efficient learning even with limited training data. A segmentation method proposed based on this idea is TextDiff \cite{Feng24}.
The TextDiff has a diffusion model-based image encoder and a Clinical BioBERT as a text encoder.
It combines embedding representations from both modalities using cross-modal attention. The resulting multi-scale attention map is upsampled and integrated to obtain the final segmentation output.
TextDiff uses pre-trained image and text encoders to improve learning efficiency while training only the cross-modal attention module and subsequent hidden layers. This enables multi-modal segmentation models to be trained effectively even with limited data.
This approach successfully developed a high-performance segmentation model using a small dataset of just 150 X-ray images.

Diffusion models are also being applied to video-based segmentation. Lu et al. proposed Diff-VPS \cite{Lu24}, a method for polyp segmentation in colonoscopy videos. This approach improves diffusion model-based segmentation accuracy by incorporating multi-task learning and temporal information.
The Diff-VPS introduces two key innovations.
One is the multi-task learning. The model solves image classification and object detection as additional subtasks within the segmentation diffusion model, enabling it to utilize high-level contextual information.
The other one is the temporal reasoning module. To incorporate temporal information from video frames, this module is trained using a task that estimates the target frame based on previous frames. This training method allows the model to extract and leverage temporal information in the diffusion-based segmentation process.
Experimental results demonstrated that Diff-VPS achieved higher segmentation accuracy than conventional methods for polyp segmentation in colonoscopy videos.

\section{Text generation AI}

\subsection{Overview}

Text generation AI is designed to generate text based on given conditions. Notable commercial services utilizing text generation AI include ChatGPT and Gemini. ChatGPT, introduced in November 2022, attracted worldwide attention because it could generate text more naturally and human-likely compared to previous text generation AIs.

This section briefly introduces natural language processing (NLP), which underlies text generation AI, and discusses LLMs, which have significantly focused on NLP research in recent years.

\subsection{Advances in NLP and emergence of LLMs}
\label{ssec:nlpllm}

NLP aims to enable computers to process natural language, which humans use daily, and solve various language-related tasks. The core tasks in NLP include text classification and generation, and research has been conducted to enhance these capabilities. Some practical applications of NLP include machine translation, kana-kanji conversion, search engines, and dialogue systems, many of which had already been studied and commercially applied even before ChatGPT emerged.

Similar to image processing, NLP models have increasingly adopted deep learning techniques. To handle sequential data such as text, early models relied on Recurrent Neural Networks (RNNs) and Long Short-Term Memory (LSTM) networks \cite{Hochreiter97}, which improved RNNs' ability to process long sequences. However, recent high-performance text generation AI models predominantly utilize Transformer-based NLP architectures \cite{Vaswani17}.
The Transformer model first converts extracted morphemes from text into embedding representations, then applies Multi-Head Self-Attention to learn the relationships between embeddings, ultimately generating the output. This mechanism has also been adapted for image processing in models such as ViT \cite{Dosovitskiy21}.

The performance of Transformer-based natural language processing models has improved significantly, leading to their widespread adoption in commercial services. OpenAI's ChatGPT uses the Generative Pre-trained Transformer (GPT), while Google's Gemini is based on PaLM 2 and a Transformer model also named Gemini. Additionally, Meta's Llama is another example of a Transformer-based model. The rapid performance gains of these NLP models are driven by the scaling up of data, model size, and computational resources.

In 2020, OpenAI published research on Scaling Laws for Neural Language Models \cite{Kaplan20}, reporting that the performance of language models improves as the amount of training data, the number of model parameters, and the computational resources used during training increase. Based on these findings, OpenAI developed the GPT models using massive datasets collected from the internet, Transformer-based models with many parameters, and extensive GPU resources.
The first GPT model (GPT-1), introduced in 2018, was trained on 4.5GB of text data, had 117 million parameters, and was trained using eight GPUs \cite{Radford18}. OpenAI rapidly expanded the scale of its models, leading to GPT-2 in 2019, which utilized 40GB of text data and had 1.5 billion parameters. GPT-3, released in 2020, was trained on 570GB of text data, had 175 billion parameters, and was trained using 10,000 GPUs (NVIDIA V100) \cite{Patterson22}. By 2022, GPT-3.5 had increased in scale to 355 billion parameters, and ChatGPT was built using a customized version of GPT-3.5. These details are summarized in Table \ref{tab:llm}.
Models with such many parameters are categorized as LLMs, which achieve high performance by leveraging vast amounts of data and large-scale computational environments.

\begin{table*}[tb]
\caption{LLMs and numbers of model parameters, scales of training data, and number of GPUs used for their development. ‘–’ indicates that information has not been disclosed} 
\label{tab:llm}
\begin{tabular}{ccccc}
\toprule
LLM & Num. of parameters & Training data & Num. of GPUs & Released \\
\midrule
GPT-1\cite{Radford18} (OpenAI) & 117M & 4.5GB (text) & 8 GPUs & June, 2018 \\
GPT-2 (OpenAI) & 1.5B & 40GB (text) & - & Feb., 2019 \\
GPT-3 (OpenAI) & 175B & 570GB (text) & 10,000 GPUs\cite{Patterson22} & June, 2020 \\
GPT-3.5 (OpenAI) & 355B & - & - & March, 2022 \\
GPT-4 (OpenAI) & - & - & - & March, 2023 \\
Llama 2 (Meta) & 70B & 2T (token) & - & July, 2023 \\
Llama 3 (Meta) & 405B & 15.6T (token) & - & July, 2024 \\
PaLM 2 (Google) & 540B & - & - & May, 2023 \\
Gemini (Google) & 1.6T & - & - & Dec., 2023 \\
\botrule
\end{tabular}
\end{table*}

Since LLMs require an enormous amount of text data for training, it is impractical to annotate correct answers for such large datasets manually. Instead, Self-Supervised Learning (SSL) is used for training.
One example of SSL in Transformer-based large-scale models is BERT \cite{Devlin19}, which is trained using a masked language modeling (MLM) approach. In this method, a sentence is converted into a fill-in-the-blank problem. For example, given the sentence ``The capital of Aichi Prefecture is Nagoya City'', the training model sees ``The capital of Aichi Prefecture is ?'' and learns to predict the missing word ``Nagoya City''.
GPT models also utilize a similar SSL approach, allowing them to learn from massive datasets without relying on manually annotated data. After the pre-training phase, the model undergoes fine-tuning and human-in-the-loop adjustments to optimize its performance for specific tasks.

\subsection{LLMs for medical applications and Japanese language}

LLMs built using general text data may not achieve sufficient performance for specialized tasks such as medical support. To address this, many LLMs trained on medical text corpora have been proposed \cite{Zhou24}. One example is BioBERT, a model with 110 million parameters trained on texts from PubMed and PMC. In addition, larger-scale models such as BioMedLM, GatorTronGPT, and PMC-LLaMA have also been introduced. These models are summarized in Table \ref{tab:llmmed}.

\begin{table*}[tb]
\caption{LLMs developed using medical text corpora. Numbers of model parameters, scales of training data, and number of GPUs used for their development are also shown. ‘–’ indicates that information has not been disclosed} 
\label{tab:llmmed}
\begin{tabular}{ccccc}
\toprule
LLM & Num. of parameters & Training data & Num. of GPUs & Released \\
\midrule
BioBERT & 110M & 18G (token) & 8 GPUs & Sep., 2019 \\
BioMedLM & 2.7B & 110GB & - & March, 2024 \\
GatorTronGPT & 20B & 277B (token) & - & Nov., 2023 \\
PMC-LLaMA & 13B & 79B (token) & 32 GPUs & Aug., 2023 \\
Med-PaLM (Google) & 540B & - & - & Dec., 2022 \\
Med-PaLM 2 (Google) & - & - & - & May, 2023 \\
\botrule
\end{tabular}
\end{table*}

LLMs built using Japanese-language corpora have also been introduced. These are summarized in Table \ref{tab:llmjp}.
ELYZA Inc. has released ELYZA-japanese-Llama-2-7b and Llama-3-ELYZA-JP-70B, versions of Meta’s Llama 2/3 models fine-tuned on Japanese text. In addition, Preferred Networks and Preferred Elements have released PLaMo-13B and PLaMo-100B, while Japan’s National Institute of Informatics (NII) has published LLM-jp-13B and LLM-jp-3 172B.
Among them, LLM-jp-3 172B is particularly notable. It has 172 billion parameters, making it comparable in scale to OpenAI’s GPT-3.
Other notable examples include LLMs fine-tuned using data from Japan’s National Examination for Medical Practitioners, such as Llama3-Preferred-MedSwallow-70B and Preferred-MedLLM-Qwen-72B, both released by Preferred Networks.

\begin{table*}[tb]
\caption{LLMs developed using Japanese text corpora. Numbers of model parameters, scales of training data, and number of GPUs used for their development are also shown. ‘–’ indicates that information has not been disclosed} 
\label{tab:llmjp}
\begin{tabular}{ccccc}
\toprule
LLM & Num. of parameters & Training data & Num. of GPUs & Released \\
\midrule
\begin{tabular}{c}
PLaMo-13B\\ (Preferred Networks)
\end{tabular}
 & 13B & 1.4T (token) & 480 GPUs & Sep., 2023 \\
\begin{tabular}{c}
PLaMo-100B\\ (Preferred Elements)
\end{tabular}
 & 100B & 2T (token) & - & Sep., 2024 \\
\begin{tabular}{c}
ELYZA-japanese-Llama-2-7b\\ (ELYZA)
\end{tabular}
 & 7B & 18B (token) & 32 GPUs & Aug., 2023 \\
\begin{tabular}{c}
Llama-3-ELYZA-JP-70B\\ (ELYZA)
\end{tabular}
 & 70B & - & - & June, 2024 \\
LLM-jp-13B (NII) & 13B & 300B (token) & 96 GPUs & Oct., 2023 \\
LLM-jp-3 172B (NII) & 172B & 2.1T (token) & - & Sep., 2024 \\
\begin{tabular}{c}
Llama3-Preferred-MedSwallow-70B\\ (Preferred Networks)
\end{tabular}
 & 70B & - & - & July, 2024 \\
\begin{tabular}{c}
Preferred-MedLLM-Qwen-72B\\ (Preferred Networks)
\end{tabular}
 & 72B & - & - & March, 2025 \\
\botrule
\end{tabular}
\end{table*}

\subsection{Medical support using LLMs}

Research on the medical applications of LLMs is rapidly advancing, enabling capabilities such as automatic generation and summarization of radiology reports, structuring medical findings, and automatic anonymization of clinical text. Numerous research results have been reported, and several companies have released LLM-based commercial medical support services.

As an example of commercial service using LLMs, Therapixel (France) offers MammoScreen \cite{MammoScreen}, which automatically detects tumors in mammogram images and generates drafts of radiology reports, including findings and impressions. In the United States, RADPAIR \cite{RADPAIR} provides tools for radiology report input via dictation and automated report generation support. In the future, more companies providing image-based diagnostic support are expected to begin incorporating LLMs to assist in radiology report creation.

Research studies using LLMs in medical support are also increasing rapidly. One of the early studies applying LLMs in this context \cite{Wang24} explored the automatic generation of descriptive text from medical images. This approach, called ChatCAD, performs tasks such as tumor detection on medical images and then uses an LLM to generate explanatory text based on the results.
Another example, introduced in Section \ref{ssec:segmentation}, is TextDiff \cite{Feng24}, which applies LLMs to medical image segmentation. This method uses Clinical BioBERT, a specialized LLM for clinical text, as a text encoder and combines textual and image information to generate segmentation results.

\section{Foundation models}

\subsection{Generalized performance of large-scale models}

In Section \ref{ssec:nlpllm} in the chapter on text generation AI, this paper introduced the Scaling Laws \cite{Kaplan20}, which state that the performance of neural language models improves with increases in the amount of training data, the number of model parameters, and the scale of computation during training. This principle applies not only to language models but also to image processing models.

A report by Zhai et al. \cite{Zhai22} demonstrated that in ViT-based image processing models, increasing the amount of training data, the number of model parameters, and the scale of computation similarly leads to performance improvements.
One well-known example of an image processing model built using massive datasets and large-scale computation is Meta’s Segment Anything Model (SAM) \cite{Kirillov23}. SAM was trained on 11 million images and over one billion region annotations, resulting in a ViT-based model with over 600 million parameters.
Another example is Flamingo by DeepMind \cite{Alayrac22}, which also leverages large-scale data and models to achieve high performance.
Meta has also released an improved version, Segment Anything Model 2 (SAM 2) \cite{Ravi24}, which enhances SAM’s accuracy and supports video processing. Compared to SAM, SAM 2 offers better accuracy and faster processing speed, incorporating temporal attention into the model. SAM 2 was trained on approximately 51,000 videos, enabling it to segment objects in video sequences accurately.

Large-scale image processing models such as SAM, sometimes called Large Vision Models (LVMs), possess general-purpose capabilities that are not limited to specific tasks. While SAM is primarily trained on natural images and performs well in segmenting natural scenes, it has also been shown to work for medical image segmentation.
Several studies have reported the applications of SAM to medical images, including tumor segmentation in pathology images \cite{Deng23}, cardiac segmentation in ultrasound images \cite{Chen23}, and segmentation across various imaging modalities \cite{Ma24}.
Although SAM’s segmentation accuracy is generally lower than that of task-specific models, it can still perform segmentation across various scenarios, regardless of differences in object scale and shape or whether the image is color or grayscale.
SAM can recognize and segment images without additional fine-tuning using the target task data (i.e., zero-shot learning), even for medical images such as pathology images. These results indicate that large-scale image processing models exhibit strong generalization capabilities.

\subsection{Foundation models and changes in AI development}

In traditional machine learning-based AI development, data collection and model development were carried out separately for each task, resulting in task-specific AI systems. However, this approach requires repeated efforts for every new task and is not efficient.

In contrast, training relatively large-scale models using cross-task datasets makes it possible to create models with general-purpose performances that can be applied to multiple tasks. These models are known as foundation models. Foundation models have the following characteristics:
\begin{itemize}
\item They are trained on large-scale, cross-task datasets.
\item They can be transferred to a wide range of downstream tasks.
\end{itemize}
SAM is a foundation model in image processing, while GPT is an example in natural language processing.

\begin{figure}[tb]
\centering
\includegraphics[width=0.8\linewidth]{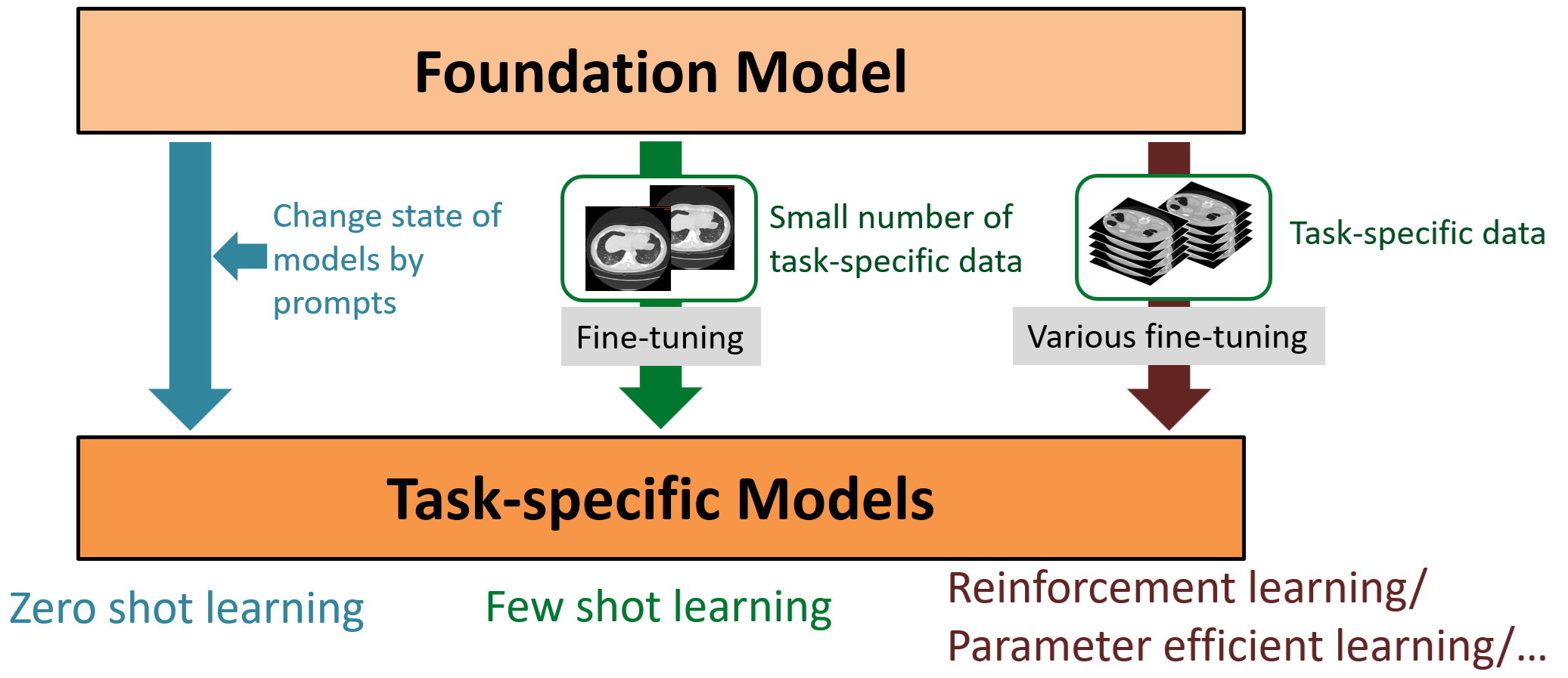}
\caption{Conceptual diagram of task-specific AI development using foundation models}
\label{fig:fm}
\end{figure}

The value of foundation models lies in their ability to reduce the burden of AI development for individual tasks. Using a foundation model minimizes the effort required for task-specific data collection and rapidly builds models with performance comparable to task-specific AI systems. This capability has the potential to transform traditional AI development methodologies.
In recent years, various approaches to developing AI for downstream tasks using foundation models have been proposed.
Figure \ref{fig:fm} illustrates some of these.
Zero-shot learning is the method of adapting a foundation model to a task without using any task-specific data by adjusting its state with prompts.
Few-shot learning is a method of fine-tuning a foundation model with a small amount of task-specific data.
Additionally, methods such as parameter-efficient learning have been proposed, which enable efficient training of large foundation models on limited computational resources.

\begin{figure}[tb]
\centering
\includegraphics[width=\linewidth]{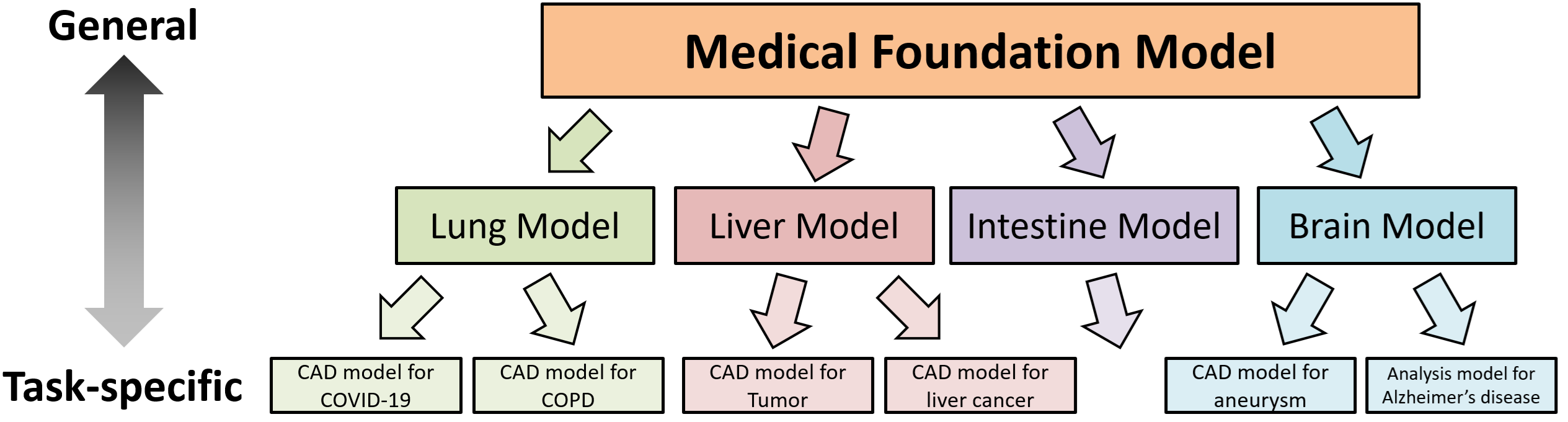}
\caption{Relationship between medical foundation model and downstream tasks as proposed by Shuo Li at SPIE Medical Imaging 2024 \cite{Li24}}
\label{fig:shuoli}
\end{figure}

Various methods of AI development using foundation models are also being explored in medical image processing. Two core challenges in medical image processing are the significant effort required to collect large-scale datasets and the limited data availability for rare diseases.
In this context, foundation models' ability to support downstream task development using only small amounts of data is extremely valuable. Further research advancements in this area are highly anticipated.
In his keynote talk at the SPIE Medical Imaging 2024 international conference, Shuo Li discussed one vision for the ideal form of a medical foundation model \cite{Li24}. His idea involved building a general-purpose medical foundation model that could be transferred to various downstream tasks in healthcare support, thereby enabling the development of AI systems for various medical tasks.
He described a hierarchical structure of downstream tasks in medical support, where tasks are organized by organ, and beneath each organ task, there are disease-specific tasks. This concept is illustrated in Fig. \ref{fig:shuoli}.

\subsection{Methods for developing vision foundation models}

This section introduces methods used to develop foundation models for image processing. In foundation model development, self-supervised learning (SSL) is typically used for pretraining, using many unlabeled images \cite{Jing20}. After pretraining, transfer learning or fine-tuning is performed using labeled images for the target downstream task, resulting in a task-specific model.
The most commonly used SSL techniques for pretraining vision foundation models include Contrastive Learning (CL), Masked Image Modeling (MIM), and a combination of CL and MIM.

Contrastive Learning (CL) generates two augmented versions of the same image through data augmentation. The model extracts feature representations using an encoder, and training is performed to increase similarity between positive pairs (images augmented from the same original image) and decrease similarity between negative pairs (images augmented from different images).
Several CL-based methods have been proposed.
Among them, SimCLR \cite{Chen20simple} is a fundamental contrastive learning method.
MoCo \cite{He20} reuses negative samples efficiently during training.
SimSiam \cite{Chen21siamese} and DINO \cite{Caron21} perform CL without using negative samples.
SimCLR2 \cite{Chen20big} and MoCov2 \cite{Chen20moco} are enhanced versions of SimCLR and MoCo, respectively.
A multimodal version of CL that utilizes image-text pairs is CLIP \cite{Radford21}, which is widely used for developing multimodal foundation models.

Masked Image Modeling (MIM) divides an image into patches, masks a portion of them, and trains the model to reconstruct the original image using the unmasked patches.
MIM is commonly used as an SSL technique when employing ViT.
A fundamental approach in MIM is the Masked Autoencoder \cite{He22mae}. It has been shown that masking up to 75\% of patches during training still leads to high accuracy in downstream tasks.
Variants such as Attention-guided MIM \cite{Kakogeorgiou22} have also been proposed, which adjust the masking process using attention weights to focus on semantically meaningful regions.

Several methods combine CL and MIM to enhance model performance.
For example, Contrastive Masked Autoencoders (CMAE) \cite{Huang23} create positive and negative image pairs using pixel shifts and then apply masking and reconstruction in the encoder similar to a masked autoencoder. The model is trained using loss functions from both CL and MIM.
Other notable approaches include ``What Do Self-Supervised Vision Transformers Learn?'' \cite{Park23}, which further explores SSL for ViT-based models.

\subsection{Foundation models for medical image processing}

One significant advantage of foundation models is their applicability across a wide range of downstream tasks. However, achieving high performance, even with transfer learning, becomes difficult if there is a large domain gap between the image data used to train the foundation model and the downstream task's image domain. To achieve better transfer learning results in specific image domains, using a foundation model built using data closely aligned with the target domain is more effective.
For example, SAM was trained on general-purpose natural images and may not provide sufficient accuracy when directly applied to medical image processing. Therefore, a variety of foundation models have been proposed that are specifically trained using single- or multi-modal medical images.
This section introduces such foundation models for medical image processing.

MedSAM \cite{Ma24} is a segmentation-oriented foundation model applicable to 2D medical images from various modalities. It was created by training the SAM architecture on medical images, including CT, MR, X-ray, ultrasound, mammography, optical coherence tomography (OCT), endoscopy, dermoscopy, fundus, and pathology images, with over 1.57 million images used for training. Unlike SAM, which allows various prompts, MedSAM uses only bounding boxes as prompts. It has been shown to outperform SAM in tasks like organ and tumor segmentation from 2D medical images.

MedSAM-2 \cite{Zhu24medsam2} extends SAM 2 to support 3D medical images and video segmentation. By treating one spatial axis of a 3D image as a temporal axis, MedSAM-2 leverages SAM 2's temporal modeling capabilities to enable 3D segmentation.

BiomedCLIP \cite{Zhang23biomedclip} is a multimodal foundation model that combines images and text. Built on the CLIP framework, it incorporates PubMedBERT \cite{Gu21pubmedbert} as its text encoder to improve performance in the medical domain. It was trained on PMC-15M, a 15 million figure-caption pairs dataset extracted from 4.4 million scientific articles in PubMed Central (PMC). BiomedCLIP supports tasks such as image classification and visual question answering (VQA).

BioViL-T \cite{Bannur23biovitl} is another multimodal model that combines X-ray images with radiology report text. It uses a custom multimodal architecture trained on about 174,100 image-text pairs from the MIMIC-CXR v2 dataset. BioViL-T can be used for image classification and text generation from images.

PathAsst \cite{Sun24pathasst} is a multimodal foundation model for pathology images and text. Based on the CLIP framework, it was trained on over 207,000 image-text pairs collected from sources such as PubMed. PathAsst is suitable for image classification and text generation.

Prov-GigaPath \cite{Xu24gigapath} is a single-modality foundation model tailored for pathology images. It was pretrained on 1.3 billion image patches extracted from 171,000 whole-slide images, using methods such as DINOv2. It is applicable to various downstream tasks in pathology.

RETFound \cite{Zhou23retfound} is a foundation model developed for ophthalmology, supporting fundus and OCT images. It was trained on 1.6 million images using the Masked Autoencoder framework. RETFound has shown excellent performance across various retinal image-based downstream tasks.

\section{Future directions in medical image processing research}

\subsection{Overview}

In general image processing and natural language processing, the development of large-scale models and foundation models has progressed by referencing the Scaling Laws. Large models such as GPTs and Gemini are developed using the vast computational resources of big tech companies, making it difficult for academic researchers to compete in model development.

In contrast, even big companies face challenges in collecting massive datasets in medical image processing. As a result, current efforts focus on developing task-specific foundation models within this domain. Moreover, in developing AI for medical support, it is essential to be mindful of data bias related to country or race. However, no foundation model that is specifically suited to the Japanese patient population has yet been realized.
To develop AI tailored to medical support needs in each country, foundation models that account for regional and demographic biases must be created. Collaboration among academic researchers in medical image processing is necessary to achieve this, particularly for data collection and model development.

What is required to make this a reality? Recalling the Scaling Laws, three key factors are necessary to improve model performance, including a large amount of training data, a large number of model parameters, and a significant scale of computational resources during training.
While preparing these elements is not easy, it is not impossible if we fully utilize national data and computational resources. Assuming the implementation of large-scale models is possible through software development, the following sections will explore how to prepare the necessary large-scale medical image datasets and computational resources.

\subsection{Building large-scale medical image datasets}

There are already examples where large-scale medical image datasets have been developed. One notable case is the "Medical Image Big Data Cloud Platform" developed by the Medical Big Data Research Center at Japan's National Institute of Informatics (NII). This platform stores over 400 million medical images across various modalities, including radiological, endoscopic, ophthalmologic, and pathology images. Since 2017, the center has collected these images through the SINET network, resulting in one of Japan's most significant medical image datasets.

Once a vast number of images is collected, the next major challenge becomes how to annotate them efficiently. A notable report by Qu et al. \cite{Qu23} describes an approach using a human-in-the-loop system to enable large-scale annotation. In their study, they annotated multi-organ abdominal segmentation regions on 8,448 CT cases (approximately 3.2 million 2D slice images) in an impressive three-week period.
Their workflow involved AI-driven automatic segmentation, followed by manual correction by human experts, which is a human-in-the-loop setup. They utilized three segmentation models, including Swin UNETR, nnU-Net, and U-Net, and leveraged model disagreement (inter-model uncertainty) to prioritize and streamline human correction. Such a system demonstrates that efficient annotation of large datasets is feasible within a reasonable time.

Additionally, synthetic and simulation data are proving highly effective in building large annotated datasets. Constructing AI models using easily generated synthetic or simulation data and then fine-tuning them on real data can significantly reduce the overall data collection cost.
Due to advances in simulation technologies, the sim-to-real approach, which is commonly used in robotics and autonomous driving, is now being actively researched in biomedical applications. One such example is the 4D eXtended CArdiac-Torso (XCAT) Phantom \cite{Segars17}, which can generate virtual medical images by simulating various biological characteristics such as age, gender, body size, organ structure, and motion on a virtual human model. It can even simulate realistic noise and imaging artifacts, making it possible to generate large-scale medical image datasets at low cost.
Another example is a study by Menten et al. \cite{Matrin22}, which used vascular development simulation for training an AI for blood vessel segmentation. They generated numerous 3D retinal vascular structures via simulation and created annotated OCT angiography images, which were used to train a segmentation AI. Their results showed that using simulated data led to higher segmentation accuracy than using only real data.
As these research efforts demonstrate, integrating synthetic and simulation data into data collection and AI development is extremely promising and valuable in the context of medical support applications.

\subsection{Utilizing large-scale computational resources}

Currently, AI development requires distributed training using many GPUs, especially for large-scale models. However, acquiring many high-performance GPUs is a significant challenge.
To overcome this, researchers can leverage supercomputers developed by universities and research institutions.
Considering the growing demands of AI development, many of these supercomputers are equipped with many GPUs and support general development environments, including Python.
With access to such infrastructure, it is entirely feasible to develop large-scale AI models.
Although using supercomputers typically incurs usage fees, several support programs are available to subsidize these costs. When effectively utilized, these programs can allow access to powerful computational resources with little to no financial burden.

As discussed throughout this chapter, large-scale medical image datasets and computational resources essential for developing foundation models for medical support are realistically available. By maximizing the use of national data and computing resources, we can anticipate the emergence of nationally developed foundation models and high-performance AI systems that contribute to the advancement of medical support technologies.

\section{Conclusions}

In this paper, the mechanisms of image generation AI and text generation AI, along with examples of their applications in the medical field are introduced. Furthermore, this paper introduced the Scaling Laws, a concept that has gained prominence as generative AI continues to evolve, and introduced foundation models, which are expected to significantly impact the future of AI development.
Finally, this paper examined the importance of preparing large-scale medical image datasets and securing large-scale computing resources to realize foundation models for medical image processing. This paper emphasized the necessity of developing foundation models for medical support using national data and computing resources.
As medical image processing enters a new era, I hope that academic researchers will continue to play a key role and make meaningful contributions to its advancement.

\backmatter

\section*{Acknowledgments}
Parts of this research were supported by the JST CREST Grant Number JPMJCR20D5, the MEXT/JSPS KAKENHI Grant Numbers 24K03262, 24H00720, and 24K15206.

\section*{Declarations}


\bmhead{Conflict of interest}
The author have no conflicts of interest regarding this manuscript.

\bmhead{Ethics approval}
Ethics approval is not necessary because no human or animal participants, specimens, or data were involved in this work.

%

\bmhead{Data availability}
No data was used in this paper.
%

\bibliography{rpt-bibliography}

\end{document}